\documentclass[anonymous=true]{article}
\usepackage{spconf,amsmath,graphicx}
\makeatletter
\def\maketag@@@#1{\hbox{\m@th\normalfont\normalsize#1}}
\makeatother
\usepackage{cite}
\usepackage[shortcuts,acronym]{glossaries}
\usepackage{graphicx}
\usepackage[skip=0pt]{caption}
\usepackage{subcaption}
\usepackage{booktabs}
\usepackage{makecell}
\usepackage{multirow}
\usepackage{hyperref}
\usepackage{todonotes}
\usepackage{svg}
\usepackage{mathtools}
\usepackage[ruled]{algorithm2e}
\usepackage{xcolor}
\everymath{\displaystyle}

\newacronym{cv}{CV}{Computer Vision}
\newacronym{bi}{BI}{Bayesian Inference}
\newacronym{dl}{DL}{Deep Learning}
\newacronym{dml}{DML}{Deep Metric Learning}
\newacronym{dnn}{DNN}{Deep Neural Network}
\newacronym{lar}{LAR}{Label-Aware Ranked}
\newacronym{ml}{ML}{Machine Learning}
\newacronym{rdis}{RDIs}{Range-Doppler Images}
\newacronym{rl}{RL}{Reinforcement Learning}
\newacronym{drl}{DRL}{Deep Reinforcement Learning}
\newacronym{ood}{OOD}{Out-of-Distribution}
\title{MEET: A Monte Carlo Exploration-Exploitation Trade-off for Buffer Sampling}
\address{Author Affiliation(s)}
\name{\begin{tabular}{c}$^{\star}$Julius Ott$^{1,2}$ \qquad $^{\star}$Lorenzo Servadei$^{1,2 }$ \qquad Jose Arjona-Medina$^{3}$ \qquad Enrico Rinaldi$^{4}$ \\ 
\qquad Gianfranco Mauro$^{1}$ \qquad Daniela Sánchez Lopera$^{1,2}$ \qquad Michael Stephan$^{1}$ \\ 
\qquad Thomas Stadelmayer$^{1}$ \qquad Avik Santra$^{1}$ \qquad Robert Wille$^{2}$\end{tabular}}

\address{$^{1}$Infineon Technologies AG,
$^{2}$ Technical University of Munich, \\
$^{3}$Johannes Kepler University Linz, 
$^{4}$University of Michigan}

\begin{document}
\setlength{\abovedisplayskip}{4pt}
\setlength{\belowdisplayskip}{4pt}
\maketitle
\begingroup\renewcommand\thefootnote{$\star$}
\footnotetext{Equal contribution}
\endgroup
\begin{abstract}
Data selection is essential for any data-based optimization technique, such as Reinforcement Learning. State-of-the-art sampling strategies for the experience replay buffer improve the performance of the Reinforcement Learning agent. However, they do not incorporate uncertainty in the Q-Value estimation. Consequently, they cannot adapt the
sampling strategies, including exploration and exploitation
of transitions, to the complexity of the task.
To address this, this paper proposes a new sampling strategy that leverages the exploration-exploitation trade-off. This is enabled by the uncertainty estimation of the Q-Value function, which guides the sampling to explore more significant transitions and, thus, learn a more efficient policy.
Experiments on classical control environments demonstrate stable results across various environments. They show that the proposed method outperforms state-of-the-art sampling strategies for dense rewards w.r.t.\ convergence and peak performance by 26\% on average.
\end{abstract}
\section{Introduction}
\label{sec:intro}

In \emph{\ac{drl}} applications, the buffer, where experiences are saved, represents a key component.
In fact, learning from stored experiences leverages supervised learning techniques, in which Deep Learning excels \cite{fedus2020revisiting}. Seminal work has shown how buffer sampling techniques improve the performance of \ac{drl} models over distributions observed during training \cite{nikolov2018information}. Consequently, how to sample from the buffer plays an important role in the learning process. In this context, a major component of the buffer sampling strategy regards the uncertainty of the agent in choosing the optimal action.
This influences the trade-off between exploration-exploitation in the buffer sampling strategy.

\noindent In the literature, the concept of uncertainty has been applied to tasks performed by a \emph{\ac{ml}} model over unseen data distributions. Those are called \emph{\ac{ood}} data, i.e., samples for which the model has high uncertainty. Thus, in the state of the art, the assessment of that uncertainty is typically used for \ac{ood} detection. For instance, \cite{uncertainty_dqn} proposes an uncertainty-based \ac{ood}-classification framework called UBOOD, which uses the epistemic uncertainty of the agent's value function to classify \ac{ood} samples. In particular, UBOOD compares two uncertainty estimation methods: dropout- and bootstrap-based. The highest performance is achieved using bootstrap-based estimators, which leverage the bootstrap neural network (BootDQN) \cite{Bootstrapped_DQN}.
\newline Inspired by \cite{Bootstrapped_DQN, uncertainty_dqn}, this paper employs a bootstrap mechanism with multiple heads for determining the uncertainty in the \mbox{Q-Value} estimation. This is exploited by the proposed novel algorithm: a \textbf{M}onte Carlo \textbf{E}xploration-
\textbf{E}xploitation \textbf{T}rade-Off (MEET) for buffer sampling. Thanks to the \mbox{Q-value} uncertainty estimation, MEET enables an optimized selection of the transitions for training Off-Policy \emph{\ac{rl}} algorithms and maximizes their return.
We evaluate MEET on continuous control problems provided by the MuJoCo \footnote{\url{github.com/deepmind/mujoco}} physics simulation engine.
Results show that MEET performs consistently in terms of convergence speed and improves the performance by 26\% in challenging, continuous control environments. \newline
The remainder of this paper is structured as follows: in Section~\ref{sec:Background}, we present the background related to continuous \ac{rl}, buffer sampling, and uncertainty estimation. Furthermore, we motivate the necessity of the proposed approach. In Section~\ref{sec:Approach}, we introduce the proposed buffer sampling strategy, while Section~\ref{sec:Experiments} describes the performed experiments on public datasets and the obtained results. Finally, Section~\ref{sec:Conclusion} concludes the paper.

\section{Background and Related Work}\label{sec:Background}
In this section, we  present concepts related to the approach introduced in this paper. To this end, we first present characteristics of continuous \ac{rl}, and then we review the role of uncertainty in \ac{rl}. 

\subsection{Continuous Reinforcement Learning}
Traditional \ac{rl} methods often assume a finite action space.
In real-world applications, however, \ac{rl} methods do face a continuous action space.
Different methods have been developed to extend existing methods to continuous action spaces.
One prominent example is the \emph{Deterministic Policy Gradient (DPG)} method \cite{silver:14}.
Using the same approach as for stochastic policies, 
the parameters are updated in the direction of the policy gradient.
In this method, exploration is achieved by learning from samples uncorrelated to the current policy.
This work was extended in Deep Deterministic Policy Gradient \cite{lillicrap:16},
by combining Deep Q-Learning and DPG,
at the expense of losing convergence guarantees of DPG.
As an alternative to the policy gradient method,
a continuous variant of the Q-Learning algorithm
(Normalized Advantage Functions)
enables the use of \mbox{Q-Learning} in continuous action spaces.
In this method, the advantage function is parameterized as a quadratic function of
nonlinear features of the state \cite{Gu:16}.
In this way, the action that maximizes the \mbox{Q-Value} function
can be determined analytically during the Q-Learning update.
Soft Actor-Critic methods \cite{Haarnoja:18}
combine Off-Policy training with a stochastic actor, which also
maximizes the entropy. 
As a result, training stability increases, in conjunction with efficiency and performance.

\noindent In desirable scenarios, optimal control is reached by having a model of the environment. Nevertheless, in many \mbox{real-world} applications, a model of the environment cannot be accurately estimated. Therefore, in those cases, \mbox{model-free} \ac{drl} is often used.
Model-free On-Policy learning is known to be very sample inefficient, as data is not reused \cite{sutton2018reinforcement}.
Sampling trajectories from the environment can be costly and slow.
To overcome this, the \ac{rl} algorithm could rely on additional 
trajectories, generated by other policies, which can be used in the learning phase.
As a matter of fact, \mbox{Off-Policy} methods can deal with trajectories sampled from different policies,
improving the sample efficiency and exploration.
One of the key elements for the success of DQN \cite{Mnih:15}
was the usage of a replay buffer or experience replay \cite{Lin:92},
where transition tuples (state, action, next-state, and reward) are stored.
This buffer then serves as a training dataset 
from which supervised learning techniques can be applied.
A naive approach would sample data evenly.
However, there might be better strategies to sample from the buffer.
For example, when transition tuples are stored
to learn a \mbox{Q-Value} function using \mbox{Q-Learning},
biased sampling towards high \emph{Temporal Difference (TD)} error tuples
can speed up the learning time \cite{Schaul:15}.
The same idea has been applied to trajectories as well \cite{Brittain:19}.
When rewards are sparse or delayed,
storing trajectories with high returns
and sampling them more often
also speeds up learning \cite{arjona-medina:19}.
A different approach, compares states along new trajectories
with the states in the buffer
to filter out those which are novel \cite{Nicholaus:22}.
Even RL has been used to learn an auxiliary policy
which samples from the replay buffer
the most useful experiences \cite{Zha:19}. \newline
Uncertainty estimation
has been used already in 
off-line \ac{rl}\cite{an2021uncertainty}, 
trajectory exploration\cite{mai2022sample} 
safe \ac{rl} \cite{Lutjens:19}, and for sampling strategies in discrete \cite{nikolov2018information} 
as well as continuous action spaces \cite{kalweit2017uncertainty}.
However, 
to the best of our knowledge,
the uncertainty of the Q-Value estimation has not yet been used to leverage exploration and exploitation for buffer sampling. In that way, the proposed method samples more useful transitions in continuous action space problems, without any further assumptions on the \ac{rl} model.
\subsection{Uncertainty-Based Reinforcement Learning}\label{subsec: Unc}

Uncertainty-based \ac{rl} aims to provide a policy with a corresponding policy estimate. A
\emph{\ac{dnn}} trained with supervised learning techniques might underperform in test when training data comes from a different distribution. \cite{peng2018sim}.
In the recent years, methods to quantify uncertainty in the predictions have been focused on image and text classification tasks \cite{NEURIPS2019_8558cb40}.
In \ac{rl}, uncertainty-based methods have been used for \ac{ood} detection. In particular, these methods focus on the DQN. The goal is to estimate how certain the agent is to choose an action as the optimal. A DQN update is determined by the current state $s_t$, action $a_t$, reward $r_t$, and next state $s_{t+1}$ in the form of $\theta_{t+1} = \theta_t + \eta(y_t^Q -Q(s_t, a_t;\theta_t)) \nabla_\theta Q(s_t,a_t;\theta)$, where $\theta$ are the network parameters, $y_t^Q$ the target value at episode step $t$, and $\eta$, the learning rate.
To estimate the uncertainty of the Q-Value function approximation, 
three methods are compared in \cite{uncertainty_dqn}: Monte-Carlo Concrete Dropout \cite{Kendall:2017}, and Bootstrap methods leveraging BootDQN \cite{Bootstrapped_DQN} with and without a random prior Network~\cite{Osband:18}. Monte Carlo Concrete Dropout employs dropout layers, which learn individual dropout rates per layer. This avoids hyperparameter search for optimal rates and couples with changing data during training, which happens in an \ac{rl} setting.

\noindent However, Monte Carlo Concrete Dropout is outperformed by bootstrap-based methods. 
Uncertainty bootstrap-based methods leverage the statistical idea of bootstrapping, i.e., using an ensemble of models to approximate a population distribution by a sample distribution. The bootstrapped DQN\cite{Bootstrapped_DQN} approximates a ``distribution" of Q-Values, not to be confused with distributional \ac{rl}. It trains $L$ estimates of the Q-Value function $\hat Q_{l}(s,a,\theta)$ against its target network $\hat Q_{l}(s,a,\theta^-)$. Two architectures are possible: an ensemble of $L$ \ac{dnn}s estimating $\hat Q_{l}$-value functions or one \ac{dnn} with $L$ heads. The multi-head approach represents a more effective method, which uses the same memory buffer over $L$ estimates and has not to be parallelized to be efficient. 
The multi-head bootstrapped DQN modifies DQN by adding $L$ heads or $L$ Q-Value functions. For each episode step $t$, a value function $\hat Q_{l}$ is selected to act by choosing $l \in \{1,...,L\}$ from a uniform and random distribution. For each step of the episode, the action $a$ maximizing $\hat Q_{l}(s_t,a)$ is executed. Afterward, a masking probability distribution $M$ generates $m_t$, a mask to identify whether the experience in $t$ should be used for training. The current transition tuple and the mask $m_t^l$ are stored in the replay buffer, which is common to all $L$ heads. The gradients $g$ of the value function $\hat Q_l$ in a time step $t$ are $g_t^l = m_t^l(y_t^Q -\hat Q_{l}(s_t, a_t;\theta)) \nabla_\theta \hat Q_{l}(s_t,a_t;\theta)$.
DQN refers to an only-critic algorithm, referring to \cite{Ac_review}. Thus, we can extend the idea of bootstrapped DQN to the Actor-Critic method without loss of generality.

\section{Constructing a Buffer Sampling Strategy based on Uncertainty}\label{sec:Approach}
In this section, we introduce MEET, an algorithm that leverages both exploration and exploitation for an improved buffer sampling strategy.
To this end, we first review how the proposed method relates to \emph{Upper Confidence Bound (UCB)} algorithms. Afterward, we explain in detail the implementation of MEET.
\subsection{A Sampling Strategy for Exploration-Exploitation Trade-off}\label{subsec:sampling_exp_exp}
Many techniques target the exploration-exploitation dilemma in \ac{rl}.
Although those techniques have been applied to reward functions, as well as to states and actions selection, they have not yet been studied for transition sampling in buffers.
Accordingly, we imagine the transition sampling as a decision-making problem, similar to the UCB algorithm in the Multi-Armed Bandit problem. Here in fact the decision-making happens, under uncertainty, by selecting one of \mbox{$k$-armed} bandits at each time step. A decision-maker or agent is present in the Multi-Armed Bandit Problem to choose between \mbox{$k$-different} actions and receives a reward based on the selected action.
In this paper, we consider UCB1, which trades-off exploitation and exploration, and is formalized as:
\begin{equation}\label{Eq:UCB1}
  \textrm{UCB1}(i) = \mu_i +  \sqrt{\frac{{2\ln{(N)}}}{{N_i}}},
\end{equation}
where $\mu_i$ represents the current reward  average of arm $i$ at the current round; $N$ the number of trials passed; and $N_i$, the number of pulls given to arm $i$ in the play through history.
Similarly, this algorithm has been applied in the literature to \mbox{tree-based} search algorithms, taking the name of Upper Confidence bound applied to Trees (UCT), introduced in \cite{kocsis2006bandit}. Accordingly, random sampling is coupled to a tree-based search algorithm for a more efficient search in a defined space using an upper confidence bound algorithm. This leads to the selection of the most promising node of the tree, from which a sequence of actions is unrolled. The confidence bound for a parent node $u$ and child node $u_i$ of a tree search is given by
\begin{equation}\label{Eq:UCT}
  \textrm{UCT}(u_i, u) = \frac{Q(u_i)}{N(u_i)} + c \frac{\ln{N(u)}}{N(u_i)},
\end{equation}
where $Q(u_i)$ corresponds to the total simulation reward for the node $u_i$, and $N(\cdot)$ identifies the number of visits to a node.
Although several versions of UCB have been proposed, as in \cite{auer2002using}, the algorithm's core takes advantage of the exploration-exploitation trade-off to find optimistic solutions for the choice of next moves in a task. \newline
Upon methods selected from exploration-exploitation strategies and Monte Carlo sampling, we introduce a novel strategy for the same problem on Replay Buffering transitions sampling, to be specific a \emph{M}onte Carlo \emph{E}xploration-\emph{E}xploitation \emph{T}rade-off (MEET) for an improved buffer sampling strategy.
Instead of sampling on a state space, MEET samples transitions in the replay buffer with an exploration-exploitation strategy, which adapts well to the confidence of the network in solving the task.
To express the mean and variance related to the exploration-exploitation strategy, we adopt the multi-head bootstrap network proposed in \cite{Bootstrapped_DQN} for estimating uncertainty in the Q-Value estimation (i.e., $\hat{Q}$). In our case, the multi-head uncertainty corresponds to the variance in the prediction of the Q-Value, as a result of the training of multiple Q-Heads. The process is described in Section \ref{subsec: Unc}. To trade-off exploration and exploitation of transitions, we elaborate a priority score $p$ for sampling a transition, i.e.,
\begin{equation}\label{Eq:Priority score}
p = \sigma^2(\hat{Q}) \left ( \mu({\hat{Q}}) + \frac{1-\mu({\hat{Q}})}{N(v)} \right),
\end{equation}
where the value of $p$ is computed for each transition stored in the buffer. \newline The number of visits for a transition $v$ is expressed by $N(v)$.
We can rewrite the same formula as: 
\begin{equation}\label{Eq:Priority score one}
p = \left(1-\frac{1}{N(v)}\right) \mu({\hat{Q}}) \sigma^2({\hat{Q}}) + \frac{1}{N(v)}  \sigma^2(\hat{Q}),
\end{equation}
where the exploitation corresponds to the first term, while the exploration is expressed with the second term . 
We notice that similarly to UCB1 and UCT bounds, both the exploitation and exploration parts consider the number of visits $N(v)$. \newline
Intuitively, if the transition has not been sampled sufficiently, the variance term will encourage exploration. The variance would be higher as the Q-Value estimation of the multiple heads disagree on the unseen sample.\newline
The more visits to the transition, the more relevant becomes the exploitation term. In this case, the multiplication of the transition's \mbox{Q-Value} mean and its variance gives a higher sampling priority score. In fact, while the Q-Value mean identifies the value of the action in that state, the variance multiplication assesses the uncertainty on the Q-Value for the given scene. Transitions with higher uncertainty and expected \mbox{Q-Value} are favored by the exploitation term, thus encouraging visits on promising scenes while multiple heads have not reached a consensus on the Q-Value estimate.

\subsection{MEET}\label{subsec:UCB_MEET}
Algorithm \ref{Alg:Meet} shows the complete pseudocode on a critic network.
\noindent During the training of the critic network, at each step of the episode, we store a transition with associated sampling score equal to the maximum among the already stored ones, according to Equation \ref{Eq:Priority score}. In this step, we make sure that new injected transitions can be explored in the sampling process.
Afterwards, as proposed in \cite{Bootstrapped_DQN}, we randomly select some critics' heads so that only a subset of them, with probability $m_p$, are trained during each epoch. These heads are called active heads and are represented as $\tilde{Q}$. During the replay period, transitions are sampled following the normalized priority score shown in Equation \ref{Eq:Priority score}. In order to compute the exploration-exploitation terms, the mean $\mu({{\hat{\tilde{Q}}}})$ and variance $\sigma^2({{\hat{\tilde{Q}}}})$ of the active heads is computed. Both are normalized for stability reasons. To use a ‘sum-tree’ implementation, the mean is also shifted by its minimum to become a positive number, which results in computational complexity of $\mathcal{O}(\log{}N)$, where $N$ identifies the number of samples in the buffer. \newline
After the computation of the critic loss, we update the active critic heads by multiplying the gradient of the loss with $1/N(v)$, to limit the bias of the transitions often sampled. \newline
In practice, the MEET algorithm can be applied to any DRL method using critic networks and can be extended to discrete action space problems without loss of generality.
Here, MEET can be used to train a critic network, and then compute the TD Loss. While only the critic network is trained using MEET, the parameters of the network are then passed to the critic target network (e.g. through Polyak Averaging \cite{lillicrap:16}). 
Those two networks contribute then to compute the TD Loss, which in turn, is used for the update step.
\setlength{\textfloatsep}{0pt}
\begin{algorithm}[ht!]
\caption{MEET Buffer Sampling}
\label{Alg:Meet}
    \SetAlgoLined
    \SetKwInOut{Input}{Input}
    \SetKwInOut{Output}{Output}
    Steps $T$. Number of critic heads $L$.
    Initialize Replay Memory $\mathbf{H} = \emptyset, \bigtriangleup = 0, p_0 = 1$, $m_p$\\
    Observe $s_0$ and choose $a_0 \sim \pi_{\theta}(s_0)$ \\
    \For{t=1 to T}{
    Observe $s_t$, $r_t$ \\
    Set the number visits: $N(v_t) = 0$ \\
    Store transition $v_t =$ ($s_{t-1}$, $a_{t-1}$, $r_t$, $s_t, N(v_t)$) \\in $\mathbf{H}$ with maximal priority $p_t = \max_{i<t}p_i$ \\
    Set Q-Head mask $m_t \sim B(1,m_p)$ \\
    Number of heads $M = \sum_{l = 1}^{L} m_t$\\
    Select active heads $\tilde{Q}$ based on $m_t$ \\
    \uIf{$t\equiv 0 \: Mod \: K$}{
    \For{$j = 1$ to $k$}{Sample transition $j \sim P(j) = p_j/ \sum_i p_i$ \\
    Compute Q-Heads mean $\mu({{\hat{\tilde{Q}}}})$ \\
    Compute Q-Heads variance  $\sigma^2({\hat{\tilde{Q}}})$  \\
    Update $N(v_j) = N(v_j) + 1$ \\
    Update priority score: \\
    $p_j = \sigma^2({\hat{\tilde{Q}}}) \left ( \mu({\hat{\tilde{Q}}}) + \frac{1-\mu({\hat{\tilde{Q}}})}{N(v_j)} \right)$ \\
    \For{m in M}{$\hat{y} = r + \gamma \cdot \hat{\tilde{Q}}_{m,target}(s_{j+1}, a_{j+1}) $ \\
    Compute Critic Loss: $J_{m,,j}$}
    }
    $J = \frac{1}{M} \sum_{m=1}^M \frac{1}{k} \sum_{j=1}^k J_{m,j} $ \\
    Accumulate weight-change $\bigtriangleup \leftarrow \bigtriangleup + \frac{1}{N(v_j)} \cdot \bigtriangledown_{\theta} J $ \\
    Update weights $\theta \leftarrow \theta + \eta \cdot \bigtriangleup$}
    Choose action $a_t \sim \pi_\theta (s_t)$
    }
\end{algorithm}
\section{Experiments}\label{sec:Experiments}
In this section, we first review the implementation settings. Afterwards, we benchmark the proposed sampling strategy against state-of-the art sampling strategies on MuJoCo, a continuous control benchmark suite.
\subsection{Implementation Settings}
In the implementation, we used PyTorch v1.8.0.\textsuperscript{\texttrademark}- GPU v2.4.0 with CUDA\textsuperscript{\textregistered} Toolkit v11.1.0. As a processing unit, we used the Nvidia\textsuperscript{\textregistered} Tesla\textsuperscript{\textregistered} P40 GPU, Intel\textsuperscript{\textregistered} Core i7-8700K CPU, and DIMM 16GB DDR4-3000 module of RAM. 
The algorithms are evaluated over nine publicly available continuous control environments. The results are averaged over five experiments per environment and the code is published on Github \footnote{\url{github.com/juliusott/uncertainty-buffer}}.
\setlength{\textfloatsep}{10pt}
\begin{figure}[htb]
\begin{subfigure}{\linewidth}
    \centering
    \includegraphics[width=\linewidth]{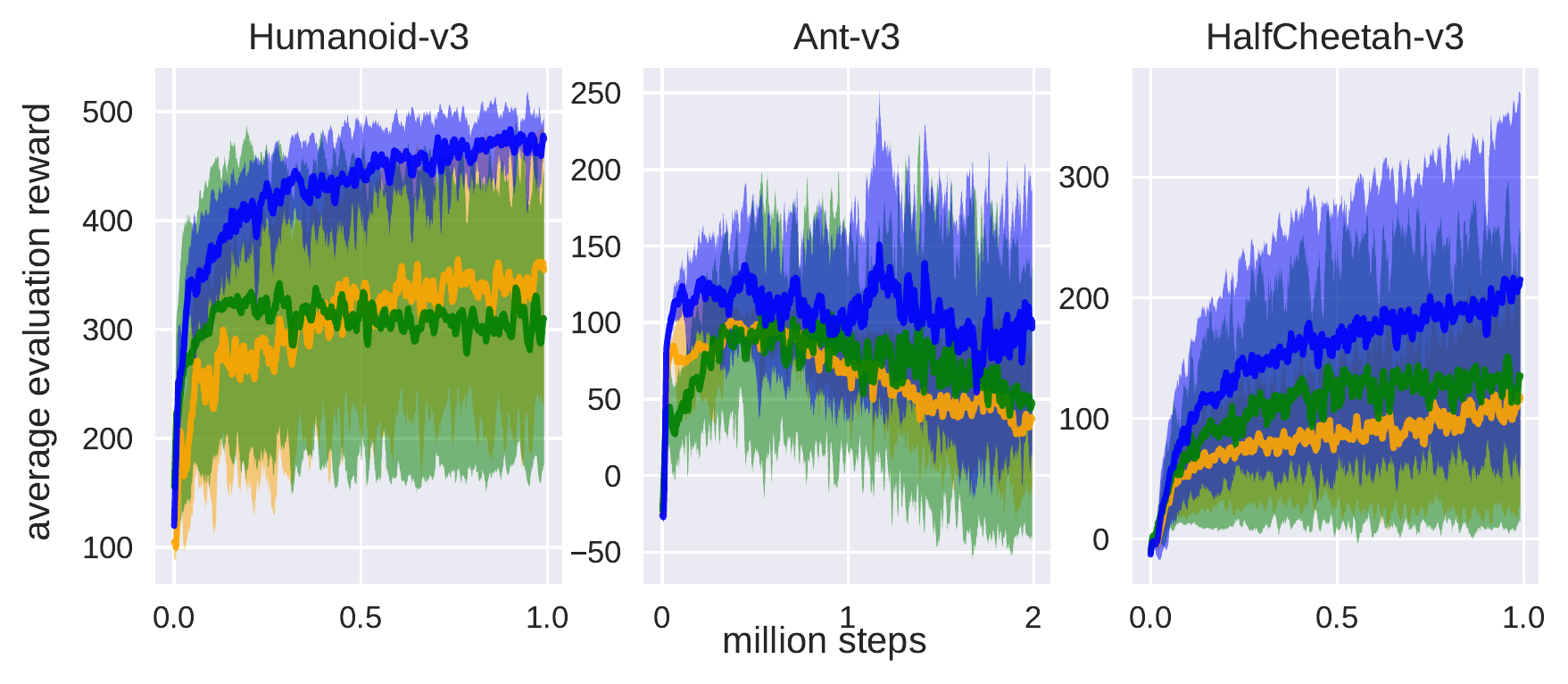}
    \caption{}
    \label{fig:mujoco_results1}
\end{subfigure}\vspace{-15pt}
\begin{subfigure}{\linewidth}
    \centering
    \includegraphics[width=\linewidth]{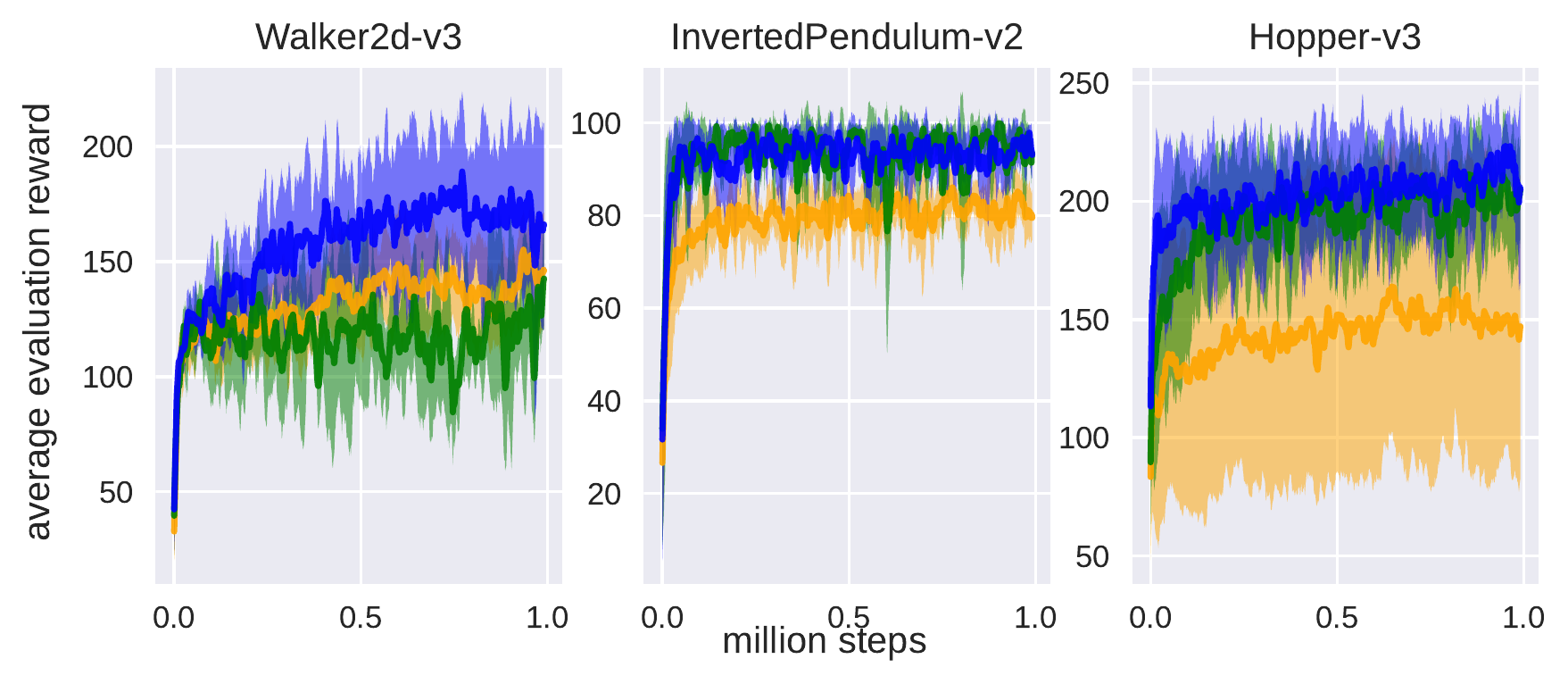}
    \caption{}
    \label{fig:mujoco_results2}
\end{subfigure}\vspace{-15pt}
\begin{subfigure}{\linewidth}
    \centering
    \includegraphics[width=\linewidth]{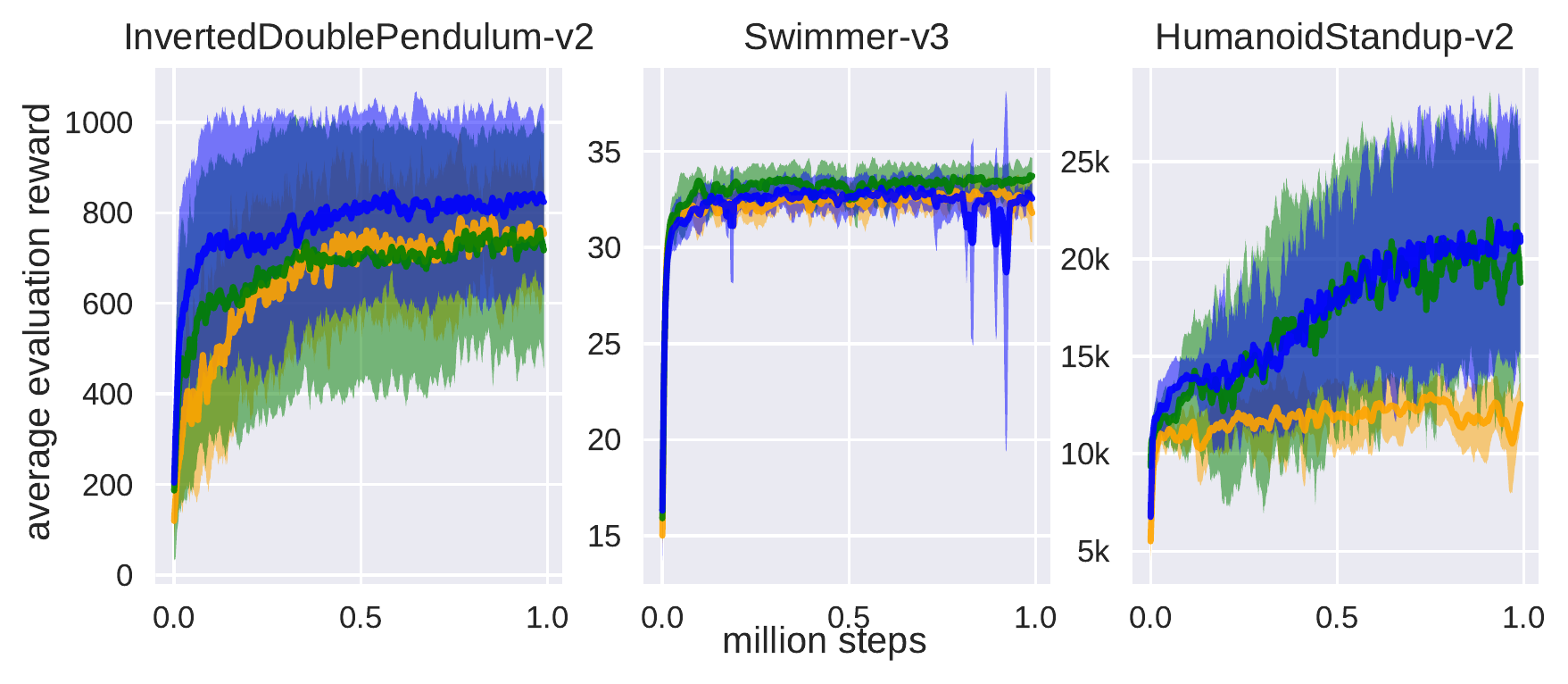}
    \caption{}
    \label{fig:mujoco_results3}
\end{subfigure}
\caption{Comparison of MEET \textcolor{blue}{(blue)}, Uniform \textcolor[rgb]{0,0.53,0}{(green)}, and Prioritized \textcolor{orange}{(orange)} Sampling Strategies on MuJoCo}
\label{fig:mujoco_results}
\end{figure}
\newline
\noindent\textbf{MEET outperforms state-of-the-art DRL buffer sampling strategies for dense rewards.} 
\noindent Figures \ref{fig:mujoco_results} illustrate the average evaluation reward attained by the soft-actor critic on different MuJoCo tasks. The MEET sampling consistently surpasses the uniform and prioritized buffer in terms of convergence and on average peak performance by 26\%.
As pointed out in \cite{barth2018distributed}, the prioritized buffer can be detrimental in continuous control problems: This characteristic is also observed in our experiments. The advantage of MEET is well underlined in the environment with the largest action space, Humanoid-v3, by the increasing performance gap. In addition, the soft-actor critic is not saturating with MEET sampling.
\section{Conclusion}\label{sec:Conclusion}
The main contribution of this paper is a method for prioritized sampling of transitions that trades off exploration-exploitation. To this end, the uncertainty estimation of the Q-Value function is used to sample more relevant transitions for the learning process. The experiments show that the presented algorithm outperforms existing methods on simulated scenarios. When benchmarked on the MuJoCo simulation environments, the MEET sampling consistently outperforms existing methods on convergence speed and performance by 26\%.
This paper evaluates MEET's performance on classical control tasks in which the action space is continuous. In future work, we expect to evaluate our approach in a discrete action space and with sparse rewards. Especially for the latter, we believe that the exploitation term is beneficial.

\bibliographystyle{IEEEbib}
\bibliography{strings,refs}

\end{document}